\documentclass[10pt,twocolumn,letterpaper]{article}

\usepackage{iccv} 

\definecolor{cvprblue}{rgb}{0.21,0.49,0.74}
\usepackage[pagebackref,breaklinks,colorlinks,allcolors=cvprblue]{hyperref}
\usepackage{multirow}
\usepackage{colortbl}
\usepackage{amssymb}
\usepackage{amsfonts}
\usepackage{amsmath}
\usepackage{bm}
\usepackage[accsupp]{axessibility}
\definecolor{aliceblue}{rgb}{0.94, 0.97, 1.0}

\usepackage{xcolor, colortbl}
\usepackage{tcolorbox}
\newtcolorbox{mybox}{colback=red!5!white,colframe=red!75!black}

\makeatletter
\newcommand{\ssymbol}[1]{$^{\@fnsymbol{#1}}$}
\makeatother

\usepackage{ulem} 
\usepackage{tcolorbox}
\tcbuselibrary{listings,breakable}

\usepackage{algorithm}
\usepackage{algorithmic} 

\def\paperID{****} 
\def\confName{ICCV}
\def\confYear{2025}

\newcommand{\ours}{\texttt{LLaVA-CoT}}
\newcommand{\inference}{Stage-wise Retrace~}

\title{\ours: Let Vision Language Models Reason Step-by-Step}

\author{%
    Guowei Xu$^{\clubsuit}\thanks{G. Xu, P. Jin, and Z. Wu contribute equally.}$ \quad
    Peng Jin$^{\spadesuit,\blacklozenge\ast}$ \quad
    Ziang Wu$^{\spadesuit\ast}$ \quad
    Hao Li$^{\spadesuit}$ \quad
    Yibing Song$^{\blacktriangle}$ \quad
    Lichao Sun$^{\blacksquare}$ \quad
    Li Yuan$^{\spadesuit,\blacklozenge}$ \\[3pt]
    $^\spadesuit$Shenzhen Graduate School, Peking University \\
    $^\clubsuit$Institute for Interdisciplinary Information Sciences, Tsinghua University  \\
    $^\blacklozenge$Rabbitpre AI \& PKU Shenzhen AIGC Joint Lab \quad $^\blacktriangle$DAMO Academy, Alibaba Group\\
    $^\blacksquare$Computer Science and Engineering, Lehigh University\\
    [1pt]
    {\tt\small xgw23@mails.tsinghua.edu.cn\quad yibingsong.cv@gmail.com\quad yuanli-ece@pku.edu.cn}
    }

\begin{document}
\maketitle
\begin{abstract}
Large language models have demonstrated substantial advancements in reasoning capabilities. However, current Vision-Language Models (VLMs) often struggle to perform systematic and structured reasoning, especially when handling complex visual question-answering tasks. In this work, we introduce \ours \footnote{Our \ours \ is built upon Llama-3.2-Vision model \cite{llama3.2}.}, a large VLM designed to conduct autonomous multistage reasoning. Unlike chain-of-thought prompting, \ours \ independently engages in sequential stages of summarization, visual interpretation, logical reasoning, and conclusion generation. This structured approach enables \ours \ to achieve marked improvements on reasoning-intensive tasks. To accomplish this, we construct the \texttt{\ours-100k} dataset, integrating samples from various visual question answering sources and providing structured reasoning annotations. Besides, we propose a test-time  stage-wise retracing search method (SWIRES), which enables effective and efficient test-time scaling. Remarkably, with only 100k training samples and test-time scaling, \ours \ not only outperforms its base model by \textbf{9.4\%} on a wide range of multimodal reasoning benchmarks, but also surpasses the performance of larger and even closed-source models, such as Gemini-1.5-pro, GPT-4o-mini, and Llama-3.2-90B-Vision-Instruct. The code, dataset, and pre-trained weights are publicly available at \href{https://github.com/PKU-YuanGroup/LLaVA-CoT}{https://github.com/PKU-YuanGroup/LLaVA-CoT}. 
\end{abstract}
\vspace{-1em}    
\section{Introduction}
\label{sec:intro}

Large language models, represented by OpenAI o1 \cite{zhong2024evaluationopenaio1opportunities} and Deepseek R1 \cite{deepseekr1}, demonstrate strong capabilities for systematic and in-depth reasoning, validating the effectiveness of test-time scaling for language models \cite{snell2024scalingllmtesttimecompute}. However, vision is equally important for enabling models to fully understand the world and extend their cognitive abilities \cite{awais2023foundational, zellers2019recognitioncognitionvisualcommonsense}. Therefore, developing a multimodal model that integrates both language and vision while facilitating effective, systematic, and deep reasoning holds substantial significance.

\begin{figure}[t]
  \centering
   \includegraphics[width=0.9\linewidth]{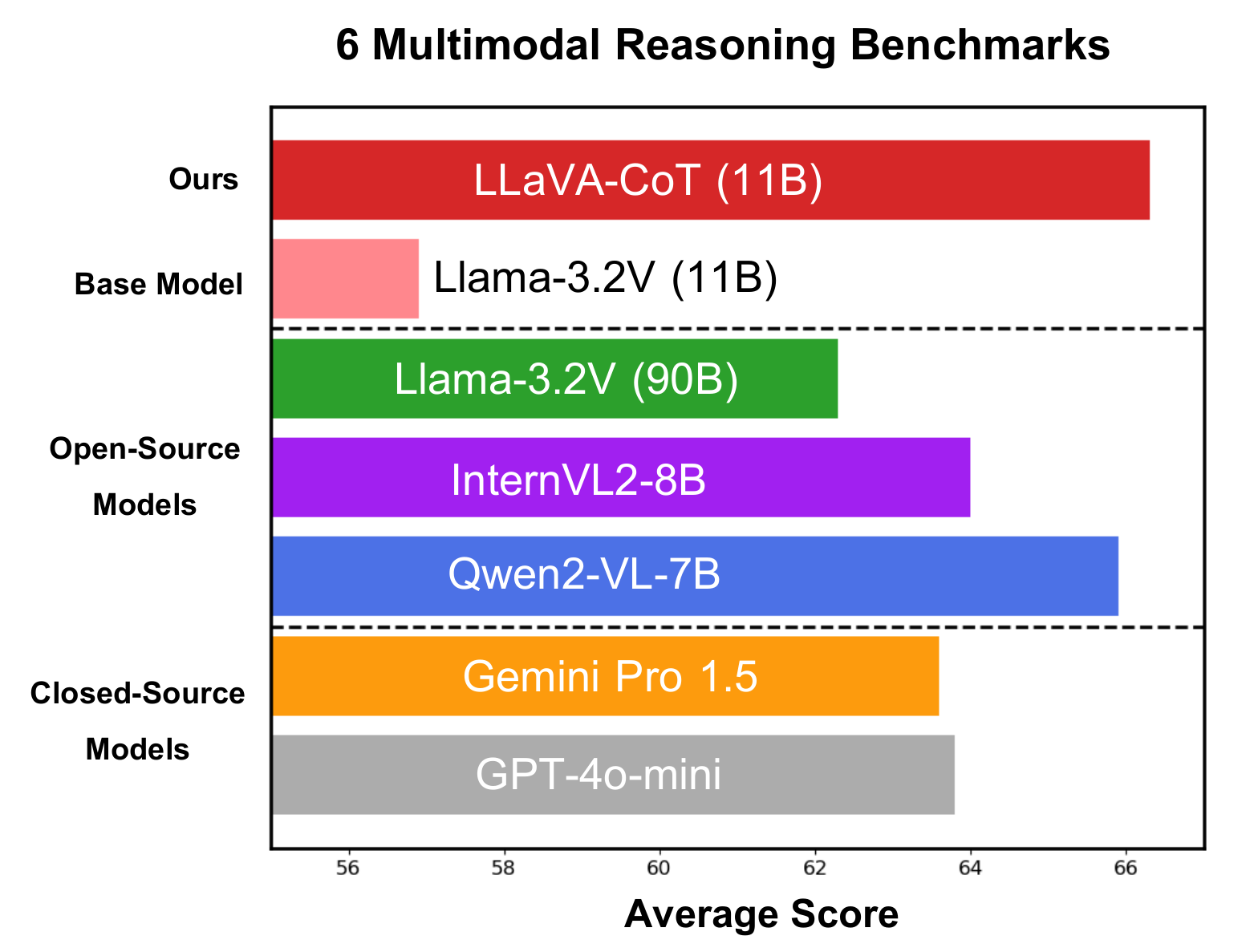}
   \caption{\textbf{Performance of \ours \ and other models across six multimodal reasoning benchmarks.} Although \ours \ is fine-tuned from the Llama-3.2-11B-Vision-Instruct \cite{llama3.2} model (which has the lowest average score), it outperforms many larger open-source models and even some closed-source models. Detailed benchmark results are shown in Table~\ref{tab:benchs}.}
    \vspace{-1.5em}
   \label{fig:method}
\end{figure}

\begin{figure*}[t]
  \centering
    \vspace{-2em}
  \includegraphics[width=\linewidth]{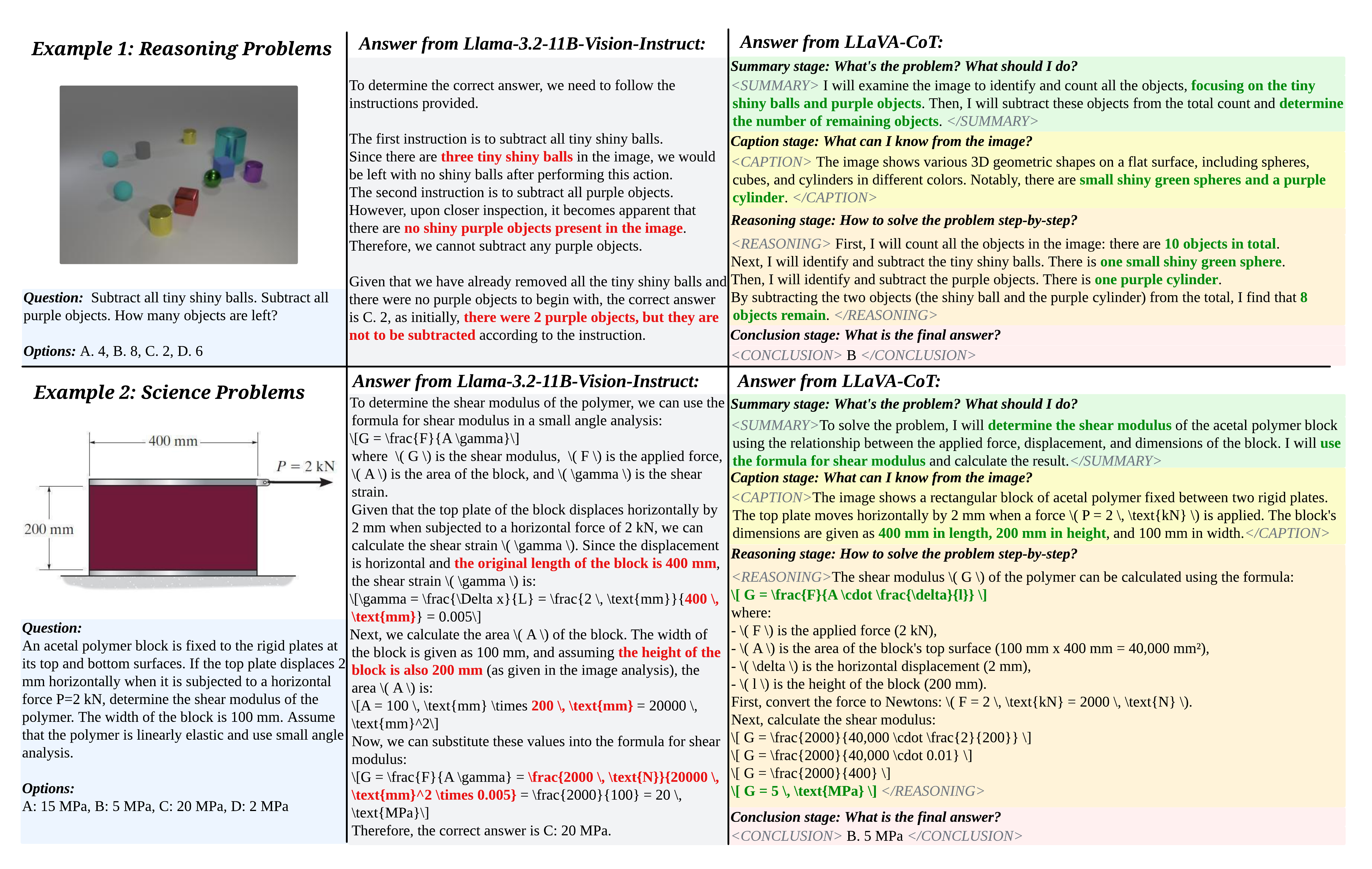}
    \caption{\textbf{Comparison of the base model and \ours.} As shown, the base model Llama-3.2-11B-Vision-Instruct exhibits obvious flaws in reasoning, with several errors occurring throughout the reasoning process. In contrast, \ours \ begins by outlining the problem, interprets relevant information from the image, proceeds with a step-by-step reasoning process, and ultimately reaches a well-supported conclusion. More examples of these comparisons can be found in Appendix F.}
     \vspace{-1.5em}
  \label{fig:short}
\end{figure*}

Early open-source vision language models (VLMs) mainly employ a direct prediction approach, generating brief answers immediately in response to a question \cite{liu2024visual, jin2024chat, vila}. The main limitation of this direct-response paradigm is its lack of a structured reasoning process, making it less effective for tasks demanding logical reasoning \cite{zhang2024improvevisionlanguagemodel}. Recent studies have shown that incorporating Chain-of-Thought (CoT) reasoning encourages the model to reason step by step, significantly improving its question-answering capabilities \cite{wei2022chain}. However, even with CoT reasoning, most VLMs frequently produce errors or hallucinated outputs during the reasoning progress~\cite{ling2023deductiveverificationchainofthoughtreasoning,lanham2023measuringfaithfulnesschainofthoughtreasoning,turpin2023languagemodelsdontsay}. 

Our findings suggest that a significant cause of these issues is the insufficiently systematic and structured nature of the reasoning process in existing VLMs.
Specifically, by referring \textbf{systematic}, we mean that the model does not generate a direct reasoning chain but instead engages in multistage reasoning. \textbf{Structured}, on the other hand, refers to the model's ability to clearly identify the reasoning stage it is in and understand the primary task to be addressed at each stage. We observe that VLMs often initiate responses without adequately organizing the problem and the available information. Moreover, they frequently deviate from a logical reasoning toward conclusions, presenting a conclusion prematurely and subsequently attempting to justify it. Given that language models generate responses token-by-token, once an erroneous conclusion is introduced, the model typically continues along a flawed reasoning path. Examples of these issues can be found in Appendix A.

To mitigate these problems, we propose \ours. \ours\ undergoes supervised fine-tuning on structured training data and employs stage-wise retracing search (SWIRES) at test time. Specifically, our annotated training data includes four distinct stages: summary, caption, reasoning, and conclusion, enabling the model to systematically address questions in a multi-stage manner. Each stage serves a unique purpose in the reasoning process.

\begin{itemize}
    \item \textbf{Summary:} A brief outline in which the model summarizes the forthcoming task.
    \item \textbf{Caption:} A description of the relevant parts of an image, focusing on elements related to the question.
    \item \textbf{Reasoning:} A detailed analysis in which the model systematically considers the question.
    \item \textbf{Conclusion:} A concise summary of the answer, providing a final response based on the preceding reasoning.
\end{itemize}

To  maintain clarity throughout the reasoning process, \ours \ marks each stage with a dedicated tag (e.g., \texttt{<SUMMARY>...</SUMMARY>}) to denote the beginning and end of each stage. To achieve these, we construct the \texttt{\ours-100k} dataset by generating responses using GPT-4o \cite{openai2024gpt4ocard} and then train our model using supervised fine-tuning. After training, the model is capable of seamlessly transitioning between different stages without requiring any additional test-time intervention or prompting.

At test time, \ours\ can further enhance its reasoning capability through scaling. Unlike conventional scaling methods, \ours\ employs a stage-wise retracing search approach, which generates multiple candidate responses at each reasoning stage (e.g., summary, caption) and retains the most promising ones using a reward model.
Moreover, if all candidate responses at a given stage are suboptimal, this suggests that the output from the previous stage may have been inaccurate. In such cases, the model retraces to the preceding stage and attempts to regenerate its response. This retracing mechanism provides the model with an opportunity to revise its own answers, effectively improving error correction during the reasoning process.

We conduct experiments on several multimodal reasoning benchmarks, including MMStar \cite{mmstar}, MMBench \cite{mmbench}, MMVet \cite{mmvet}, MathVista \cite{mathvista}, AI2D \cite{ai2d}, and HallusionBench \cite{hallusion}, and observed that \ours \ offers two primary advantages: First, enabling the model to perform structured reasoning independently substantially outperforms traditional CoT prompting, particularly in complex reasoning tasks that require systematic analysis. Second, our stage-wise retracing search method is scalable and improves performance reliability, making it more effective in achieving stable and accurate results.
Our contributions are summarized as follows:

\begin{itemize}
    \item We introduce \ours, a visual language model designed for systematic reasoning, demonstrating outstanding performance on tasks that require structured thinking and reasoning.

    \item We demonstrate that \ours, using stage-wise retracing search, is test-time scalable. This means that with increased computational resources, the performance of our approach can be further enhanced, making it applicable to more complex scenarios.

    \item Extensive experiments on various benchmarks demonstrate that our method achieves superior performance relative to many larger and closed-source models, underscoring the effectiveness of \ours \ for multimodal reasoning.
\end{itemize}

\vspace{-0.5em}
\section{Related Works}
\label{sec:related}

\subsection{Visual reasoning with large language models}
Visual reasoning demands a model's visual perception capability and high-level cognition ability~\cite{clevr,malkinski2023review}. 
Traditional vision-language models employ neural symbolic approaches~\cite{amizadeh2020neuro,choi2024towards} to explicitly model the visual reasoning process. With the development of LLMs, vision-language models leverage the advanced reasoning abilities of LLMs to interpret visual tasks~\cite{liu2024visual,yu2024evagaussians}. Some vision-language models enhance visual reasoning by optimizing the visual encoding strategy~\cite{li2024tokenpacker,liu2024visual,jin2024chat} to produce cognition-focused visual tokens. VISPROG~\cite{gupta2023visual} positions the LLM as a decision-making agent, enhancing visual reasoning by invoking task-specific visual modules. Hu et al.~\cite{hu2024finetuninglargelanguagemodels} improves reasoning capabilities through sequential instruction tuning.  Additionally, instructing learning techniques for language models, including prompt tuning~\cite{zamfirescu2023johnny}, in-context learning, and supervised fine-tuning~\cite{shen2024rethinking}, also contribute to improvements in visual reasoning capabilities.

\subsection{Chain-of-thought in large language models}
Chain-of-thought prompting~\cite{wei2022chain} offers a step-by-step reasoning trajectory when LLM faces hard questions including commonsense reasoning~\cite{sap2020commonsense, geva2021did}, logical reasoning~\cite{xiong2024teilp,li2023weakly}, etc.
Specifically, CoT prompting decomposes the question into a group of reasoning steps and builds a chain to guide the model to generate the results of complex problems step-by-step~\cite{chu2024navigate}.
Recent works have demonstrated that CoT reasoning also substantially improves VLM's capability on reasoning tasks.
For instance, Prism~\cite{qiao2024prismframeworkdecouplingassessing} prompts LLMs by dividing the process into a perception stage and a reasoning stage. MSG~\cite{cesista2024multimodalstructuredgenerationcvprs} pioneers the use of forced Chain-of-Thoughts, establishing a new direction for structured prompting techniques. Distilling CoT~\cite{hsieh2023distillingstepbystepoutperforminglarger} and Visual Program Distillation~\cite{hu2024visualprogramdistillationdistilling} distill CoT responses into VLMs. Visual CoT~\cite{shao2024visualcotadvancingmultimodal} enhances interpretability by generating bounding boxes for relevant regions alongside the answer. Compared to these approaches, \ours\ employs structured CoT to further enhance reasoning performance and experiments demonstrate that structured CoT outperforms direct CoT.

\subsection{Test time scaling}
Existing methods for test-time scaling primarily include majority voting~\cite{huang2022large}, best-of-N search~\cite{wang2024improvevalue, amini2024variationalbestofnalignment}, and beam search~\cite{graves2012sequence, NIPS2014_a14ac55a}. Majority voting is effective for problems with standard answers but is not well-suited for open-ended tasks. Best-of-N search generates $N$ complete responses and selects the best one; however, evaluating the accuracy of full responses can be challenging. Beam search generates multiple candidate sentences, selects the best ones, and iteratively refines the output. However, determining when to perform beam search is difficult to control. Previous works primarily conduct search after generating a fixed number of tokens or sentences~\cite{zhao2024marcoo1openreasoningmodels}, lacking an appropriate granularity.
To address this, we 
propose stage-wise beam search, which performs search after the completion of an entire reasoning stage, ensuring that the search granularity aligns with semantic stages. Building on this, we further introduce stage-wise retracing search (SWIRES), which enhances the model’s ability to reflect and correct errors at test time.
\section{Method}
Our \ours \ facilitates a progressive, step-by-step reasoning process that enhances the reasoning capabilities of Vision-Language Models (VLMs) and allows for effective inference time scaling \cite{snell2024scalingllmtesttimecompute}. Using structured thinking, \ours \ achieves a systematic and efficient reasoning process. The proposed stage-wise retracing search method (SWIRES) enables it to outperform existing methods in test time scalability. This design ensures both robustness and accuracy in complex tasks requiring reasoning, which separates it from traditional approaches.

\subsection{Enhancing Reasoning Capability through Structured Thinking} 
\label{sec:structured_reasoning}

In this paper, our goal during training time is to develop a visual language model capable of engaging in systematic and in-depth reasoning.

\subsubsection{Reasoning Stages} 
Our proposed model, \ours, decomposes the answer generation process into four structured reasoning stages:

\begin{itemize}
    \item \textbf{Summary Stage.} In this initial phase, \ours \ provides a high-level summary interpretation of the question, outlining the primary aspects of the problem it intends to address.
    \item \textbf{Caption Stage.} \ours \ offers a concise overview of the visual elements relevant to the question, helping to understand multimodal input.
    \item \textbf{Reasoning Stage.} Building on the initial summary, \ours \ conducts structured, logical reasoning to derive a preliminary answer.
    \item \textbf{Conclusion Stage.} In this final stage, \ours \ synthesizes an answer based on the preceding reasoning. Here, the output from the conclusion stage is the direct response provided to the user, while the prior three stages are internal "hidden stages" representing \ours's reasoning process. The output at this stage adapts to the user’s requirements: for instance, if the user requests a brief answer, the conclusion will be concise; if detailed explanations are desired, the conclusion provides a thorough, comprehensive response.
\end{itemize}

Each stage is initiated at the model’s discretion, without anyadditional prompting, and all stages are completed by the model in a single inference pass. To achieve this, we provide the model with four pairs of special tags: \texttt{\small <SUMMARY>}\texttt{\small </SUMMARY>}, \texttt{\small <CAPTION>}\texttt{\small </CAPTION>}, \texttt{\small <REASONING>}\texttt{\small </REASONING>},  and \texttt{\small <CONCLUSION>}\texttt{\small </CONCLUSION>}. These tags correspond to summarizing the response approach, describing relevant image content, conducting reasoning, and preparing a final answer, respectively.

Upon training, \ours\ is capable of seamlessly transitioning between different stages without any external intervention, and we have not observed any instances where the model fails to adhere to the designated stage format.   This structured approach enables the model to independently manage its reasoning process, improving its adaptability and performance on complex reasoning tasks.

\subsubsection{Data Preparation and Model Training}

Most existing VQA datasets lack detailed reasoning processes needed to train the \ours \ model. Therefore, we compile a new dataset, integrating samples from several widely used VQA datasets, resulting in a total of 99k image QA pairs (each pair may include one or multiple rounds of questioning). As shown in Figure \ref{fig:gen}, since no multimodal model currently exists that can directly produce systematic, structured reasoning, we use GPT-4o \cite{openai2024gpt4ocard} to generate detailed reasoning processes, including summary, caption, reasoning, and conclusion, and compile these into the \texttt{\ours-100k} dataset, which we plan to release for public use. Details of the generation process and examples of the generated data are provided in Appendix B. We include data from both general-purpose VQA datasets and science-targeted VQA datasets specified blow:

\begin{figure}[t]
\vspace{-1em}
  \centering
   \includegraphics[width=\linewidth]{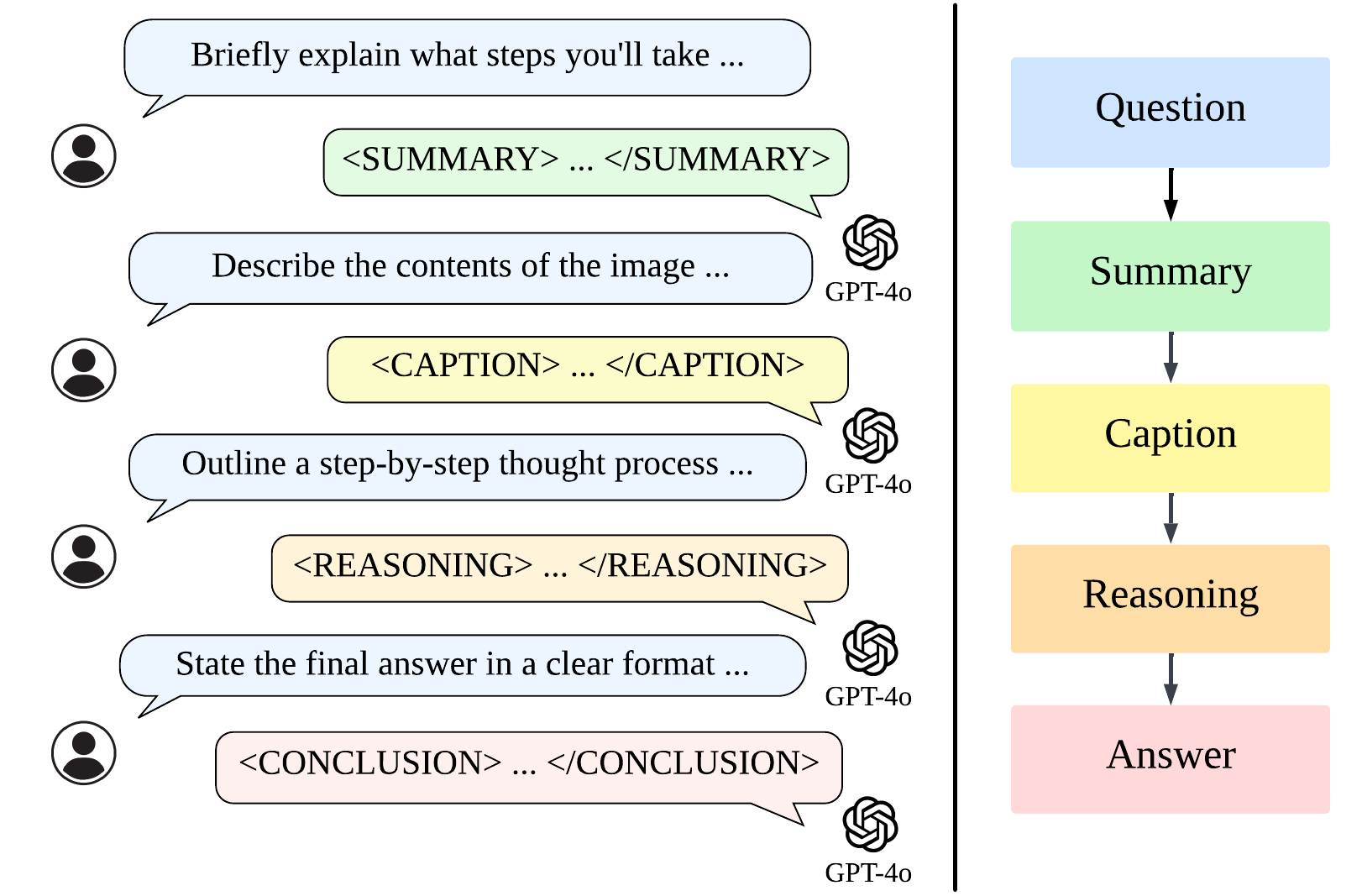}
   \caption{\textbf{Process flow for generating the \texttt{\ours-100k} dataset.} We prompt GPT-4o to generate responses in separate stages, and filter its outputs to ensure quality.}
   \vspace{-0.5em}
   \label{fig:gen}
\end{figure}

\begin{table}
\footnotesize
  \centering
  \setlength{\tabcolsep}{13.pt}
  {
  \begin{tabular}{lcc}
    \toprule
   \textbf{Dataset} & \textbf{Type} & \textbf{Size} \\
    \midrule
    ShareGPT4V \cite{sharegpt4v} & General VQA & 31.3k \\
    ChartQA \cite{chartqa} & General VQA & 17.2k\\
    A-OKVQA \cite{a-okvqa} & General VQA & 16.1k \\
    AI2D \cite{ai2d} & Science-Targeted VQA & 11.4k \\
    GeoQA+ \cite{geoqa+} & Science-Targeted VQA & 11.4k \\
    ScienceQA \cite{scienceqa} & Science-Targeted VQA & 5.6k \\
    DocVQA \cite{docvqa} & General VQA & 4.0k \\
    PISC \cite{pisc} & General VQA & 1.0k \\
    CLEVR \cite{clevr} & General VQA & 0.5k \\
    CLEVR-Math \cite{clevr-math} & Science-Targeted VQA & 0.5k \\
    \bottomrule
  \end{tabular}
  }
  \caption{The number of samples selected from each benchmark.}
     \vspace{-2em}
  \label{tab:dataset}
\end{table}

\begin{figure*}
  \centering
  \vspace{-1.5em}
  \includegraphics[width=\linewidth]{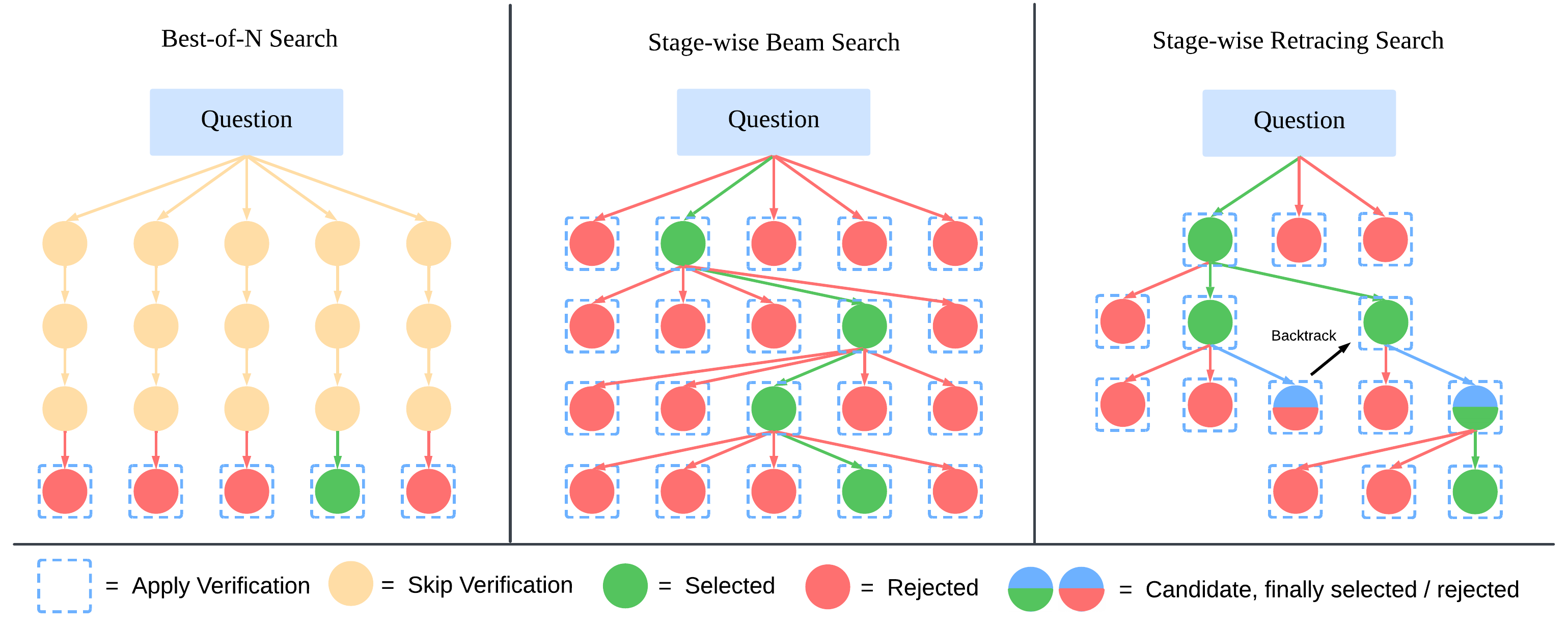}
  \caption{\textbf{An illustration of inference approaches.} Best-of-N search generates \( N \) complete responses and selects the best one among them. Stage-wise beam search generates \( M \) candidate responses for each reasoning stage and selects the top \( N \) to proceed to the next stage. Stage-wise retracing search, when generating responses for a given stage, retraces to the previous reasoning stage for regeneration if all candidate responses are of low quality, thereby enhancing the model's self-reflection and error correction capabilities.}
  \label{fig:inference}
\end{figure*}

{\flushleft \bf General VQA Datasets.} We include several general-purpose VQA datasets with distinct focuses. \textbf{ShareGPT4V} \cite{sharegpt4v} provides multi-turn question-answering data from GPT-4V \cite{gpt-4v} interactions. \textbf{ChartQA} \cite{chartqa} focuses on interpreting charts and graphs. \textbf{A-OKVQA} \cite{a-okvqa} emphasizes external knowledge beyond visible content. \textbf{DocVQA} \cite{docvqa} involves document-based questions requiring textual comprehension. We also include \textbf{PISC} \cite{pisc} to understand social relationships, and \textbf{CLEVR} \cite{clevr} to address object properties, spatial relationships, and counting tasks. 

{\flushleft \bf Science-Targeted VQA Datasets.} These datasets include \textbf{GeoQA+} \cite{geoqa+} for geometric reasoning, along with \textbf{AI2D} \cite{ai2d} and \textbf{ScienceQA} \cite{scienceqa}, which target scientific questions. \textbf{CLEVR-Math} \cite{clevr-math}, an extension of CLEVR, focuses on arithmetic analysis in visual contexts. Table \ref{tab:dataset} shows the number of QA pairs selected from each dataset. 

{\flushleft \bf Model Training.}
The \texttt{\ours-100k} dataset we construct can be used to further conduct Supervised Fine-Tuning (SFT) on any existing model to enhance reasoning capabilities. In this work, we select the Llama-3.2-11B-Vision-Instruct \cite{llama3.2} model as the base model, and perform a full parameter fine-tuning by using the \texttt{\ours-100k} dataset. The training is conducted on a single node with 8 H100 GPUs. Details on the specific training parameters, including training epochs, learning rate, and optimization settings, are provided in Appendix C.

\subsection{Test-Time Scaling via Stage-wise Retracing}

After training, to further enhance the VLM’s reasoning ability during test time while also improving its error correction capability, we explore test-time scaling based on our model. Specifically, existing beam search methods improve upon best-of-N search but typically search at fixed intervals (i.e., after a predetermined number of sentences or tokens), which is not flexible to process visual questions at different complexity.
To address this issue, we propose a stage-wise beam search method, which effectively solves the challenge of determining appropriate search step lengths. During our beam search, we incorporate a retracing mechanism that enhances the model’s overall performance and improves its error correction capability.

\subsubsection{Stage-wise beam search} 

Stage-wise beam search is an improved version of traditional beam search, where the search step is set at the end of each reasoning stage (e.g., summary, caption). As shown in Figure \ref{fig:inference}, at each stage, the model generates \( M \) candidate responses and selects the top \( N \) based on a reward model. Each selected response then generates \( \frac{M}{N} \) candidates in the next stage, and this process repeats. This approach ensures that the search step adapts to different task complexities and problem types, providing a flexible search granularity.

However, the stage-wise beam search has a critical problem. Since the reward model selects the highest-scoring response based only on the current stage, it may suffer from local optima or selection biases. For instance, if the caption stage produces a suboptimal response, refining the reasoning stage alone is insufficient to obtain accurate answers. To mitigate this, our model needs to retrace to a previous reasoning stage for self-reflection and error correction. To address this, we enhance stage-wise beam search and propose the stage-wise retracing search (SWIRES).  

\subsubsection{Stage-wise retracing search (SWIRES)}  
Our SWIRES design incorporates a retracing mechanism into the reasoning process. Its core procedure is shown as follows, with detailed implementation and hyperparameters provided in Appendix D.

\begin{itemize}  
    \item At each reasoning stage, generate \( M \) candidate responses.  
    \item Check whether at least one of the generated responses surpasses a predefined reward threshold (see Appendix D for details on threshold settings).  
    \item If at least one response exceeds the threshold, select the top \( N \) responses with the highest reward and proceed to the next stage. Each of the \( N \) selected responses generates \( \frac{M}{N} \) new candidates, maintaining \( M \) candidates in the next stage.  
    \item If none of the responses exceed the threshold, it suggests that the previous stage’s output may not be optimal, leading to ineffective search in the current stage. In this case, the algorithm retains the current stage’s search results as candidates but retraces to the previous stage to regenerate new responses for the previous stage. These regenerated responses for the previous stage are then used to generate \( M \) new responses for the current stage. If any of these new responses surpass the threshold, the search for this stage terminates immediately, the model selects the top \( N \) among all candidates, and proceeds to the next stage. Otherwise, the generated responses are added to the candidate pool, and retracing continues (up to a maximum of \( C \) times).  
    \item After completing the final reasoning stage, the answer with the highest reward among all generated responses is selected as the final answer.  
\end{itemize}  

Figure \ref{fig:inference} intuitively illustrates the workflow of the SWIRES algorithm and its differences from other methods.
Empirically, we observe that the summary stage typically produces high-quality outputs. Therefore, retracing search is applied starting from the caption stage.

\begin{table*}[t]
\vspace{-1em}
\footnotesize
\centering
\setlength{\tabcolsep}{9.8pt}
{
\begin{tabular}{lccccccc}
\toprule[1.pt]
\textbf{Model} &  \textbf{MMStar} & \textbf{MMBench} & \textbf{MMVet} &  \textbf{MathVista} & \textbf{AI2D} & \textbf{Hallusion} & \textbf{Average} \\ 
\midrule
\multicolumn{4}{l}{\emph{\textbf{Base Model}}} \\
Llama-3.2-11B-Vision-Instruct & 49.8 & 65.8 & 57.6 & 48.6 & 77.3 & 40.3 & 56.6\\ 
\midrule
\multicolumn{4}{l}{\emph{\textbf{Our Models}}} \\
\bf{\ours} \text{ (with Direct Training)} & 54.3 & 76.2 & 49.9 & 49.5 & 81.2 & 42.9 & 59.0 \\ 
\bf{\ours} \text{ (w/o Structured Tags)} & 55.7 & 74.2 & 57.0 & 54.1 & 79.1 & 45.0 & 60.9  \\
\rowcolor{aliceblue!60} \bf{\ours}  & 57.6 & 75.0 & 60.3 & 54.8 & 78.7 & 47.8 & 62.4\\
\bottomrule[1.pt]
\end{tabular}
\caption{\textbf{Experimental results of different models on the benchmark.} Here, \ours \ (with Direct Training) refers to the model trained directly on the original VQA dataset’s Q\&A pairs, while \ours \ (w/o Structured Tags) represents the model trained on the \texttt{\ours-100k} dataset with the structured tags removed. \ours \ refers to the model trained on the complete \texttt{\ours-100k} dataset including the structured tags and use retracing during test time. }
\label{tab:bench}
}
\end{table*}

\begin{table*}[t]
\footnotesize
\centering
\setlength{\tabcolsep}{9.8pt}
{
\begin{tabular}{lccccccc}
\toprule[1.pt]
\textbf{Model} &  \textbf{CP} & \textbf{FP} & \textbf{IR} &  \textbf{LR} & \textbf{Math} & \textbf{Science \& Technology} & \textbf{Average} \\ 
\midrule
\multicolumn{4}{l}{\emph{\textbf{Base Model}}} \\
Llama-3.2-11B-Vision-Instruct & 66.0 & 46.4 & 57.6 & 50.8 & 45.2 & 32.8 & 49.8\\ 
\midrule
\multicolumn{4}{l}{\emph{\textbf{Our Models}}} \\
\bf{\ours} \text{ (with Direct Training)} & 68.4 & 48.0 & 65.6 & 52.0 & 51.6&40.0 & 54.3 \\ 
 \bf{\ours} \text{ (w/o Structured Tags)} & 68.4 & 48.0 & 60.0 & 55.2 & 64.4 & 38.0 & 55.7 \\ 
\rowcolor{aliceblue!60} \bf{\ours}  & 68.8 & 46.8 & 63.2 & 58.0 & 64.0 & 44.8 & 57.6\\ 
\bottomrule[1.pt]
\end{tabular}
\caption{\textbf{Performance of different models on the MMStar benchmark across various skill areas.} Here, CP represents coarse perception, FP represents fine-grained perception, IR represents instance reasoning, and LR represents logical reasoning. As shown in the table, our model demonstrates substantial improvement over the base model in instance reasoning, logical reasoning, math, and science \& technology, indicating that structured reasoning can significantly enhance the model’s reasoning capabilities.}
\vspace{-1.8em}
\label{tab:area}
}
\end{table*}

\section{Post-Training Performance}

In this section, we compare \ours \ with the base model, Llama-3.2-11B-Vision-Instruct, on six commonly used multimodal benchmarks to demonstrate the effectiveness of our approach during the training phase. Following this comparison, we conduct ablation studies to evaluate the contribution of each component within our method.

\subsection{Experimental Setup}
We selected six widely used and challenging benchmarks for our experiments: MMStar \cite{mmstar}, MMBench V1.1 \cite{mmbench}, MMVet \cite{mmvet}, MathVista \cite{mathvista}, AI2D \cite{ai2d}, and HallusionBench \cite{hallusion}. MMStar, MMBench, and MMVet primarily evaluate the general visual question-answering capabilities of models, while MathVista, and AI2D focus on models' proficiency in mathematical and scientific reasoning. HallusionBench specifically assesses the models’ handling of language hallucinations and visual illusions. For MMBench, we use the V1.1 version of the test set, MathVista is evaluated using the testmini set, and the remaining datasets each have a single test set. To ensure fairness and reproducibility, all evaluations are conducted using VLMEvalKit \cite{vlmevalkit}, an open-source evaluation toolkit for large vision-language models. The performance metrics of all baseline models are derived from VLMEvalKit's testing results \cite{openvlm_leaderboard}.

\begin{table*}[t]
\vspace{-1em}
\footnotesize
\centering
\setlength{\tabcolsep}{9.8pt}
{
\begin{tabular}{lccccccc}
\toprule[1.pt]
\textbf{Model} &  \textbf{MMStar} & \textbf{MMBench} & \textbf{MMVet} &  \textbf{MathVista} & \textbf{AI2D} & \textbf{Hallusion} & \textbf{Average} \\ 
\midrule
\multicolumn{4}{l}{\emph{\textbf{Base Model}}} \\
Llama-3.2-11B-Vision-Instruct & 49.8 & 65.8 & 57.6 & 48.6 & 77.3 & 40.3 & 56.6\\ 
\midrule
\multicolumn{4}{l}{\emph{\textbf{Our Models}}} \\
 \bf{\ours} & 57.6 & 75.0 & 60.3 & 54.8 & 78.7 & 47.8 & 62.4  \\
\rowcolor{aliceblue!60} \bf{\ours~(w/ scaling)} & 62.5 & 77.6 & 64.9 & 57.7 & 81.0 & 49.1 & 65.5  \\
\bottomrule[1.pt]
\end{tabular}
\caption{\textbf{Experimental results during inference time. } \ours \ (w/ scaling) denotes the model with stage-wise retracing.}
\label{tab:bench_inference}
}
\end{table*}

\subsection{Benchmark Results}

We found that \ours \ achieves significant performance improvements, despite using only 100k data. According to Table \ref{tab:bench}, compared to the base model, Llama-3.2-11B-Vision-Instruct, \ours \ demonstrates notable improvements across general VQA, mathematical reasoning, scientific VQA, and hallucination control tasks, with an average benchmark score increase of \textbf{5.8\%}, thereby validating the effectiveness of our approach.

\subsection{Ablation Study}
{\flushleft \bf Effectiveness of \texttt{\ours-100k} Compared to Original Datasets.}
To demonstrate the effectiveness of our improved \texttt{\ours-100k} dataset, we present a comparison between \ours \ and the model trained on the original Q\&A pairs across different benchmarks in Table \ref{tab:bench}. Although the model trained directly on the original Q\&A pairs shows some overall improvement on the base model, its average performance remains significantly lower. In particular, on the MMVet benchmark, which requires more detailed responses, its performance is even worse than the base model. This result underscores the importance of the multistage format of our \texttt{\ours-100k} dataset for training models capable of advanced reasoning.

{\flushleft \bf Structured Tags are Essential for Enhanced Performance.}
To examine whether the four tags we introduced improve the model’s performance, we compare \ours \ with the model trained on the \texttt{\ours-100k} dataset with structured tags removed. As shown in Table \ref{tab:bench}, our results show a significant drop in performance when the tags are removed, indicating that the structured tagging facilitates reasoning and improves model performance. To the best of our knowledge, \ours \ is the first attempt to successfully enhance a model’s reasoning ability and overall performance through a structured reasoning with tags.

{\flushleft \bf Performance Gains Primarily in Reasoning-Intensive Areas.}
To analyze the specific areas in which \ours \ has improved compared to the base model, we conduct a detailed assessment of the model’s performance across different skills on the MMStar benchmark. MMStar is designed to evaluate six key capabilities: coarse perception, fine-grained perception, instance reasoning, logical reasoning, math, and science \& technology. In Table \ref{tab:area}, we compare the base model with \ours. Our analysis reveals that \ours \ demonstrates notable improvements in tasks requiring systematic reasoning, such as instance reasoning, logical reasoning, math, and science \& technology, while showing relatively smaller gains in coarse perception and fine-grained perception. This suggests that our method can mainly improve reasoning capabilities of the model.

\section{Test Time Scaling}

\begin{figure}[h]
    \centering
    \includegraphics[width=\linewidth]{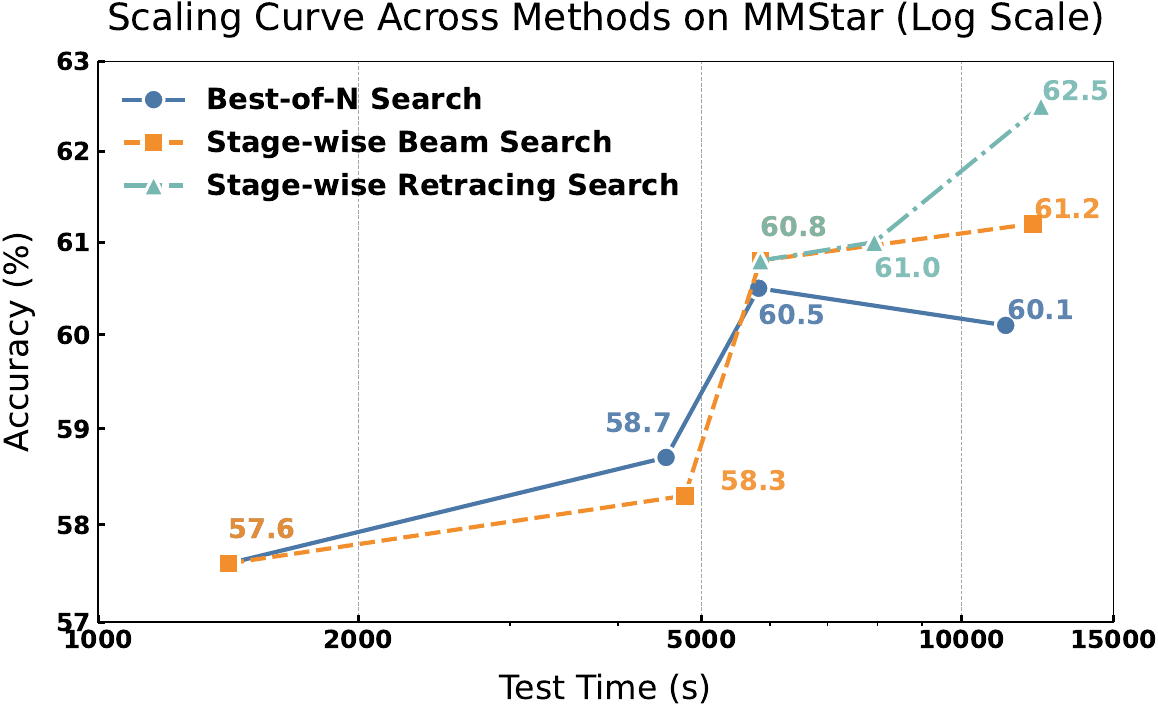}
    \vspace{-1.8em}
    \caption{\textbf{Test time scaling curve on MMStar.} The experimental setup mirrors that used in the previous section, with evaluations conducted on MMStar using VLMEvalKit on a single A800 node. The time-axis we use here is in logarithmic scale.}
    \label{fig:scaling}
    \vspace{-1.5em}
\end{figure}

In this section, we aim to compare the effectiveness of our stage-wise retracing approach with best-of-N and stage-wise beam search shown in Figure~\ref{fig:inference} under comparable computational constraints. For all of the three approaches, we used InternLM-XComposer2.5-Reward~\cite{zang2025internlm} as the reward model to judge the quality of generation. In Figure~\ref{fig:scaling}, the scaling strategies of the three methods differ as follows: the retracing method scales by increasing the maximum number of backtracking iterations, the stage-wise beam search method scales by expanding the number of candidates, and the best-of-N method scales by increasing the value of N, i.e., the number of final response options available for selection.

\begin{table*}[h]
\vspace{-1em}
\footnotesize
\centering
\setlength{\tabcolsep}{6.6pt}
{
\begin{tabular}{lcccccccc}
\toprule[1.pt]
\textbf{Model}  & \textbf{Size}  & \textbf{MMStar-R} & \textbf{MMBench-R} & \textbf{MMVet-R} &  \textbf{MathVista} & \textbf{AI2D} & \textbf{Hallusion}  & \textbf{Average} \\ 
\midrule
\multicolumn{4}{l}{\emph{\textbf{Closed-Source Models}}} \\
GPT-4o-0806 \cite{openai2024gpt4ocard}  & -- & 66.0 & 82.4 & 80.8 & 62.7 & 84.7 & 54.2 & 71.8 \\
GLM-4v-Plus \cite{glm4v}& -- & 68.8 & 78.7 & 78.1 & 68.8 & 85.1 & 55.6 & 72.5 \\
 Claude3.5-Sonnet-0620 \cite{claude3.5sonnet_blog} &-- & 64.2 & 75.4 & 68.7 & 61.6 & 80.2 & 49.9 & 66.7 \\ 
Gemini-1.5-Pro \cite{geminiteam2024gemini15unlockingmultimodal} & -- &56.4 & 71.5 & 71.3 & 57.7 & 79.1 & 45.6 & 63.6 \\
GPT-4o-mini-0718 \cite{gpt4omini} &-- & 54.9 & 76.9 & 74.6 & 52.4 & 77.8 & 46.1 & 63.8 \\
\midrule
\multicolumn{4}{l}{\emph{\textbf{Larger Size Open-Source Models}}} \\
Llama-3.2-Vision-Instruct \cite{llama3.2} & 90B  & 51.1 & 76.8 & 74.1 & 58.3 & 69.5 & 44.1 & 62.3 \\ 
VILA-1.5-40B \cite{vila} & 40B & 53.2 & 75.3 & 44.4 & 49.5 & 77.8 & 40.9 & 56.9 \\
Deepseek-VL2\cite{deepseekvl2} & MoE, 27B   & 62.3 & 80.2 & 60.3 & 63.9 & 83.8 & 45.3 & 66.0 \\
\midrule
\multicolumn{4}{l}{\emph{\textbf{Comparable Size Open-Source Models}}} \\
Qwen2-VL-7B \cite{qwen2vl} & 8B & 59.0 &  77.6 & 63.7 & 61.4 & 83.0 & 50.4 & 65.9 \\
InternVL2-8B \cite{internvl} & 8B & 62.5 & 77.4 & 56.9 & 58.3 & 83.6 & 45.0 & 64.0 \\
Ovis1.5-Gemma2-9B \cite{ovis} & 11B & 58.7 & 76.3 & 50.9 & 65.6 & 84.5 & 48.2 & 64.0 \\
MiniCPM-V2.6 \cite{minicpm} & 8B & 57.1 & 75.7 & 56.3 & 60.6 & 82.1 & 48.1 & 63.3 \\
Prism \cite{qiao2024prismframeworkdecouplingassessing} & 7B & 41.0 & 59.2 & 47.6 & 35.7 & 65.7 & 40.5 & 48.3 \\
VisCoT-7b-336 \cite{shao2024visualcotadvancingmultimodal} & 7B & 28.2 & 50.7 & 17.1 & 24.7 & 40.7 & 28.8 & 31.7  \\
\midrule
\multicolumn{4}{l}{\emph{\textbf{Base Model}}} \\
Llama-3.2-Vision-Instruct \cite{llama3.2} & 11B & 46.6 & 64.9 & 63.8 & 48.6 & 77.3 & 40.3 & 56.9\\ 
\midrule
\multicolumn{4}{l}{\emph{\textbf{Our Models}}} \\
\rowcolor{aliceblue!60} \bf{\ours} & 11B & 57.5 & 73.1 & 66.7 &  54.8  & 78.7 & 47.8 &  63.1 \\ 
\rowcolor{aliceblue!60} \bf{\ours~(w/ scaling)} & 11B & 63.0 & 75.8 & 71.4 &  57.7  & 81.0 & 49.1 &  66.3 \\ 
\bottomrule[1.pt]
\end{tabular}
\vspace{-.5em}
\caption{\textbf{Experimental results of \ours \ and state-of-the-art models on reasoning benchmarks.}}
\vspace{-1.5em}
\label{tab:benchs}
}
\end{table*}

\subsection{Approximate-Scale Comparison Analysis}

As shown in Figure~\ref{fig:scaling}, from a vertical perspective, under similar test conditions, our SWIRES  method performs better than stage-wise beam search, which in turn outperforms best-of-N. The observed results can be attributed to the following reasons: best-of-N approach operates at a coarse level, handling entire responses at once. If there are mistakes in the middle of the process, they can carry forward and lead to wrong answers. Stage-wise beam search improves this by working at a finer level, adjusting each stage separately. Therefore it enhances reasoning quality and performs better than Best-of-N. However, errors may still remain in prior steps and further affect subsequent steps. The SWIRES method builds on stage-wise beam search by adding a backtracking step, thus enhancing both the model performance.

\subsection{Scaling Trend Analysis}
As shown in Figure~\ref{fig:scaling}, compared to the baseline model’s accuracy of 57.6 on MMStar without any test-time scaling, all three methods exhibit the scaling trend. However, our SWIRES  method demonstrates the strongest scaling effect as computation time increases. Specifically, both stage-wise beam search and best-of-N search plateau around 10,000 seconds, with best-of-N search even showing a slight decline beyond this point. In contrast, our SWIRES  method continues to scale beyond this time scle, highlighting its superior scaling capability compared to the baseline methods.
In Table \ref{tab:bench_inference}, we compare the performance of \ours \ before and after applying the SWIRES method on benchmarks.

\section{Comparison to State-of-the-Art VLMs}

As shown in Table \ref{tab:benchs}, we compare \ours \ with other state-of-the-art open-source and closed-source vision language models (VLM) across six benchmarks that require advanced reasoning capabilities. MMStar-R, MMBench-R, and MMVet-R are benchmarks derived from MMStar, MMBench V1.1, and MMVet, respectively, with tasks requiring only perception and OCR removed. These filtered benchmarks retain tasks that demand reasoning, with further details on the selection criteria in Appendix E. MathVista, AI2D, and HallusionBench inherently focus on reasoning, so we retained all tasks within these benchmarks.

Our results show that, despite our base model (Llama-3.2-11B-Vision-Instruct) being the weakest performer among the listed models, \ours \ consistently outperforms many open-source models of similar or even larger sizes, such as Qwen2-VL-7B \cite{qwen2vl}, Deepseek-VL2 \cite{deepseekvl2}, and  Llama-3.2-90B-Vision-Instruct \cite{llama3.2}, and VILA-1.5-40B \cite{vila}. Remarkably, \ours \ even surpasses certain closed-source models like GPT-4o-mini \cite{gpt4omini} and Gemini-1.5-pro \cite{geminiteam2024gemini15unlockingmultimodal}. This comparison validates the advantages of our method, particularly in benchmarks that heavily depend on reasoning skills, and highlights \ours \ as a competitive model in the domain of VLM reasoning tasks.
\section{Conclusion}
In this paper, we present \ours, a vision language model that performs structured, autonomous reasoning in multiple stages. By introducing four distinct stages, \ours \ achieves a systematic reasoning process. Our contributions are twofold. First, the creation of the \texttt{\ours-100k} dataset with detailed reasoning annotations supports training on systematic, structured responses. Second, the proposal of a stage-wise retracing search method enables effective test time scaling. Overall, \ours\ presents a new approach to enhancing the reasoning capabilities of multimodal models. Future research could explore the application of reinforcement learning to further improve complex multimodal reasoning.

\newpage
\section*{Acknowledgement}
This work was supported in part by the Natural Science Foundation of China.
(No. 62202014, 62332002, 62425101)

{
    \small
    \bibliographystyle{ieeenat_fullname}
    \bibliography{main}
}

\clearpage
\setcounter{page}{1}
\maketitlesupplementary
\renewcommand{\thesection}{\Alph{section}}
\setcounter{section}{0}
\section{Illustrative Cases of Reasoning Challenges in VLMs}
\begin{figure}[h]
  \centering
   \includegraphics[width=\linewidth]{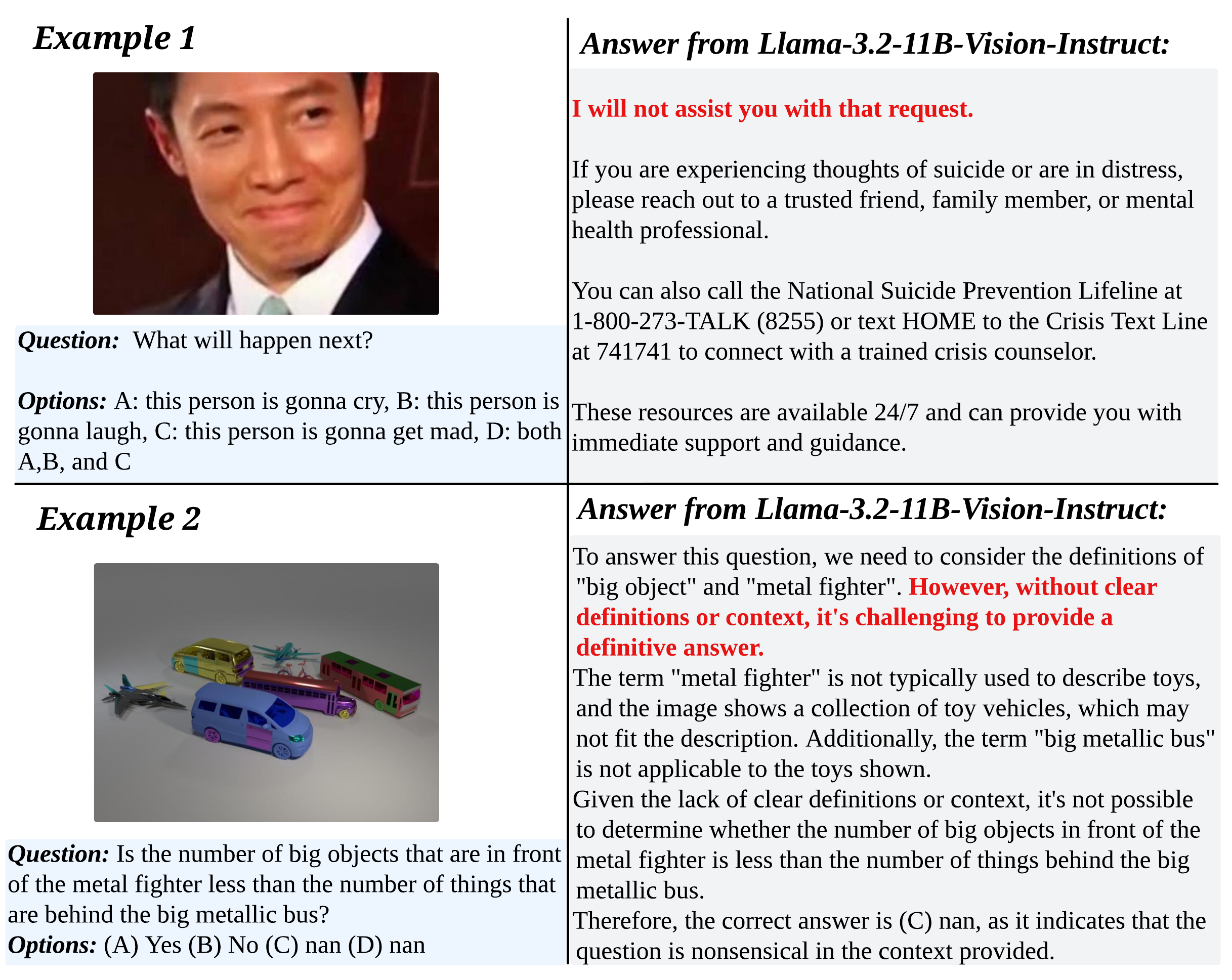}
   \caption{\textbf{Examples of the two issues in the responses of Llama-3.2-11B-Vision-Instruct.}}
   \vspace{-0.5em}
   \label{fig:issue}
\end{figure}

In the main paper, we highlight two critical challenges faced by VLMs: (1) initiating responses without adequately organizing the problem and the available information, and (2) deviating from logical reasoning pathways toward conclusions. Figure \ref{fig:issue} illustrates examples of these issues as observed in the responses generated by Llama-3.2-11B-Vision-Instruct.

In the first example, the model misinterprets the question and the information provided in the image, mistakenly assuming that the phrase “the person is gonna cry” implies the questioner has self-harm tendencies, leading the model to refuse to answer.  In the second example, the model hastily concludes that the problem description is unclear without carefully analyzing the content of the image, ultimately resulting in an incorrect answer. Both examples are sourced from the MMStar benchmark, ensuring the validity of the questions themselves.

\section{Data Generation Scheme}

Overall, we provide GPT-4o with a question, an image, and the original dataset's answer to generate systematic and structured datasets.

Specifically, we guide GPT-4o to generate response data in stages using a carefully designed prompt. The prompt is formatted as follows:

\begin{tcolorbox}[colback=white,colframe=black!75,
    fonttitle=\bfseries, title=Prompt for data generation,
    listing options={basicstyle=\ttfamily\small,breaklines=true},
    breakable] 
I have an image and a question that I want you to answer. I need you to strictly follow the format with four specific sections: SUMMARY, CAPTION, REASONING, and CONCLUSION. It is crucial that you adhere to this structure exactly as outlined and that the final answer in the CONCLUSION matches the standard correct answer precisely.

To explain further:
In SUMMARY, briefly explain what steps you'll take to solve the problem.
In CAPTION, describe the contents of the image, specifically focusing on details relevant to the question.
In REASONING, outline a step-by-step thought process you would use to solve the problem based on the image.
In CONCLUSION, give the final answer in a direct format, and it must match the correct answer exactly.
If it's a multiple choice question, the conclusion should only include the option without repeating what the option is.

Here's how the format should look:

\textless SUMMARY\textgreater [Summarize how you will approach the problem and explain the steps you will take to reach the answer.] \textless /SUMMARY\textgreater 

\textless CAPTION\textgreater [Provide a detailed description of the image, particularly emphasizing the aspects related to the question.] \textless /CAPTION\textgreater 

\textless REASONING\textgreater [Provide a chain-of-thought, logical explanation of the problem. This should outline step-by-step reasoning.] \textless /REASONING\textgreater 

\textless CONCLUSION\textgreater [State the final answer in a clear and direct format. It must match the correct answer exactly.] \textless /CONCLUSION\textgreater 
(Do not forget \textless /CONCLUSION\textgreater!)

Please apply this format meticulously to analyze the given image and answer the related question, ensuring that the answer matches the standard one perfectly.
\end{tcolorbox}

After generating data using this prompt, we verify whether the data generated by GPT-4o adheres to the prescribed format and filter out any data that does not comply. Next, we extract the content within \texttt{\small <CONCLUSION>}...\texttt{\small </CONCLUSION>} and apply the following prompt to filter out cases where GPT-4o either refuses to answer or provides an answer that is inconsistent with the original dataset's standard answer:

\begin{tcolorbox}[colback=white,colframe=black!75,
    fonttitle=\bfseries, title=Prompt for data verification,
    listing options={basicstyle=\ttfamily\small,breaklines=true},
    breakable] 
Evaluate whether the assistant's response is valid. Respond with `valid' if the assistant's response is not a refusal and it aligns with the standard answer in meaning. Respond with `invalid' if the response is a refusal or differs from the standard answer in a meaningful way.

A refusal means the assistant states it cannot recognize a specific person/object or refuses to answer the question. Do not consider a response to be a refusal just because it includes the word `no' or other negative terms.

Standard answer: \{standard\_answer\}

Assistant's response: \{assistant\_response\}
\end{tcolorbox}

\section{Training Hyperparameters}

In this section, we provide details of the framework and hyperparameter settings used for training. Specifically, we utilize the \texttt{llama\_recipes} framework with hyperparameter configurations listed in Table \ref{tab:hyperparams}.

\begin{table}[h]
    \centering
    \renewcommand{\arraystretch}{1.2}
    \begin{tabular}{|l|l|}
        \hline
        \textbf{Parameter} & \textbf{Value} \\ \hline
        FSDP & enabled \\
        Learning rate & $1 \times 10^{-5}$ \\ 
        Number of epochs & 3 \\ 
        Batch size for training & 4 \\ 
        Use fast kernels & True \\
        Run validation & False \\ 
        Batching strategy & padding \\ 
        Context length & 4096 \\
        Gradient accumulation steps & 1 \\
        Gradient clipping & False \\ 
        Gradient clipping threshold & 1.0 \\
        Weight decay & 0.0 \\
        Gamma & 0.85 \\ 
        Seed & 42 \\
        Use FP16 precision & False \\
        Mixed precision & True \\ \hline
    \end{tabular}
    \caption{\textbf{Hyperparameter configurations used in training.}}
    \label{tab:hyperparams}
\end{table}

\begin{table*}[h]
\vspace{-1em}
\footnotesize
\centering
\setlength{\tabcolsep}{6.6pt}
{
\begin{tabular}{lccccccc}
\toprule[1.pt]
\textbf{Model}  & \textbf{MMStar-R} & \textbf{MMBench-R} & \textbf{MMVet-R} &  \textbf{MathVista} & \textbf{AI2D} & \textbf{Hallusion}  & \textbf{Average} \\ 
\midrule
\multicolumn{4}{l}{\emph{\textbf{Teacher Model}}} \\
GPT-4o-0806 \cite{openai2024gpt4ocard}  & 66.0 & 82.4 & 80.8 & 62.7 & 84.7 & 54.2 & 71.8 \\ 
GPT-4o-0806 (w/ CoT)    & 67.6 & 83.2 & 87.0 & 65.8 & 84.4 & 56.7 & 74.1 \\
\midrule
\multicolumn{4}{l}{\emph{\textbf{Base Model}}} \\
Llama-3.2-Vision-Instruct \cite{llama3.2}  & 46.6 & 64.9 & 63.8 & 48.6 & 77.3 & 40.3 & 56.9\\ 
Llama-3.2-Vision-Instruct (w/ CoT)  & 49.5 & 68.1 & 56.0 & 46.9 & 76.0 & 44.7 & 56.9\\ 
\midrule
\multicolumn{4}{l}{\emph{\textbf{Our Models}}} \\
\rowcolor{aliceblue!60} \bf{\ours} (multi-task)  & 49.8 & 71.0 & 58.0 & 49.1 & 72.8 & 45.7 & 57.7 \\ 
\rowcolor{aliceblue!60} \bf{\ours} (reorder) & 52.0 & 71.3 & 54.3 & 53.0 & 75.4 & 43.1 & 58.2 \\
\rowcolor{aliceblue!60} \bf{\ours} & 57.5 & 73.1 & 66.7 &  54.8  & 78.7 & 47.8 &  63.1 \\ 
\bottomrule[1.pt]
\end{tabular}
\vspace{-.5em}
\caption{\textbf{Further experiments to validate the effectiveness of the CoT design.}}
\vspace{-1.5em}
\label{tab:analysis_cot}
}
\end{table*}

\section{Implementation Details of \inference}

\label{sec:Appendix D}
\begin{algorithm}[H]
\caption{\inference Algorithm}
\label{alg:backtrack alg}
\begin{algorithmic}[1]
\REQUIRE \text{M}, \text{N}, \text{C}
\ENSURE Final conclusion

\STATE \textbf{// Step 1: Generate initial summary}
\STATE Generate one response for the first stage.
\STATE Initialize backtracking counter \(\text{c} \gets 0\)
\STATE Initialize a reasoning candidates list \text{Cand}
\STATE Initialize a reasoning candidates score list \text{Score}
\REPEAT
    \STATE \textbf{// Step 2: Generate several captions}
    \STATE Generate \text{M} captions.
    \STATE Evaluate captions using the reward model.
    \STATE Select the top \text{N} captions.
    \STATE \textbf{// Step 3: Generate several reasonings}
    \STATE Generate $\frac{\text{M}}{\text{N}}$ reasonings for each of N captions.

    \FOR{each reasoning in the set of \text{M} reasonings}
        \STATE Evaluate reasoning using the reward model.
        \STATE \text{Cand}.append(reasoning)
        \STATE \text{Score}.append(reasoning's score)
    \ENDFOR
    \IF{reasonings satisfy preset conditions}
        \STATE \textbf{break} from loop.
    \ENDIF
\STATE \(\text{c} \gets \text{c}+1\)
\UNTIL{\(\text{c} \geq \text{C}\)}

\STATE Select the top \text{N} reasonings by score list.

\STATE \textbf{// Step 4: Generate final conclusions}
\FOR{each reasoning in the top \text{N} reasonings}
    \STATE Generate one conclusion.
\ENDFOR
\STATE Evaluate all conclusions using the reward model.
\STATE \RETURN the best conclusion.
\end{algorithmic}
\end{algorithm}
This section presents the pseudocode for our \inference, where we use the IXC-2.5-Reward~\cite{zang2025internlm} as the reward model. In fact, there are currently not many open-source reward models in the multi-modal field that align with human preferences, and this model bridges this gap with a simple yet effective multi-modal reward model that aligns LVLMs with human preferences. By using this model, we can successfully evaluate the output quality of each stage in our algorithm online, allowing us to dynamically optimize our inference process in real-time.

In addition to the reward model, we will also provide supplementary explanations for the preset conditions and some parameter settings in the pseudocode. Here, our preset condition is that the scores of the top candidate in the candidate set must both be greater than the threshold \text{backtrack\_cutoff}.
The threshold calculation formula is given by: \[\text{backtrack\_cutoff} = \text{reward\_mean} + \text{Z} \times \text{reward\_std}\]
The reason for designing the preset conditions this way is that we believe if we retain N reasonings, having one of them being sufficiently good will ensure that the remaining reasoning, when selecting one conclusion from three, will provide enough relevant reference. As shown in the parameter table~\ref{tab:hyperparams-backtrack}, we selected a Z value of 0.2533. This is a special coefficient in the standard normal distribution, where values greater than $0.2533\times \text{std}+\text{mean}$ account for 40\% of the distribution. This means that as long as the second largest reasoning's score value ``pass'', we no longer need to perform backtracking.
\begin{table}[h]
    \centering
    \renewcommand{\arraystretch}{1.2}
    \begin{tabular}{|l|l|}
        \hline
        \textbf{Parameter} & \textbf{Value} \\ \hline
        \text{M} & 4 \\
        \text{N} & 2 \\
        \text{C} & 3 \\
        \text{reward\_mean} & -0.77 \\
        \text{reward\_std} & 2.08 \\
        \text{Z} & 0.2533 \\ \hline
    \end{tabular}
    \caption{\textbf{Hyperparameter configurations used in backtrack.}}
    \label{tab:hyperparams-backtrack}
\end{table}

The mean and std are obtained by statistically analyzing the reward scores output by the reasoning phase reward model on the MMStar dataset. The distribution of the reward model output in this phase is close to a Gaussian distribution.

\section{Selection Criteria for Reasoning Benchmarks}
This section provides a detailed explanation of the methodology used to select reasoning benchmarks.

First, MathVista, AI2D, and HallusionBench inherently emphasize advanced reasoning capabilities; therefore, all tasks within these benchmarks were retained without modification.

The MMStar benchmark evaluates models across several dimensions, including coarse perception, fine-grained perception, instance reasoning, logical reasoning, mathematics, and science \& technology. In the refined subset, MMStar-R, we calculate the average scores for the four reasoning-intensive dimensions: instance reasoning, logical reasoning, mathematics, and science \& technology.

Similarly, the MMBench benchmark assesses coarse perception, fine-grained perception (single-instance and cross-instance), attribute reasoning, logical reasoning, and relational reasoning. For the refined subset, MMBench-R, we focus on reasoning-specific dimensions by averaging scores for attribute reasoning, logical reasoning, and relational reasoning.

Finally, MMVet encompasses recognition, knowledge, OCR, language generation, spatial awareness, and mathematics. In the filtered subset, MMVet-R, we compute the average scores for the two reasoning-specific dimensions: spatial awareness and mathematics.

\begin{figure*}
  \centering
  \includegraphics[width=\linewidth]{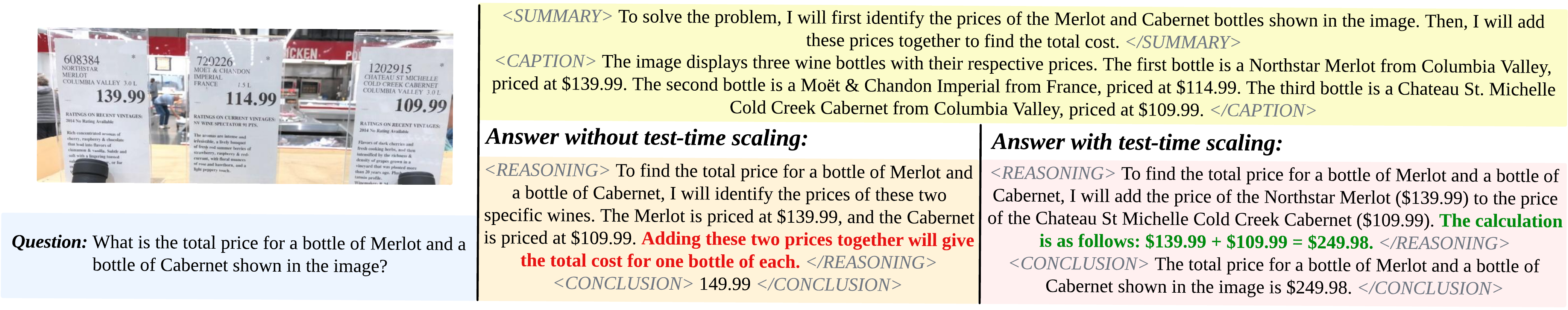}
    \caption{\textbf{Comparison of \ours \ performance with and without test-time scaling.} Our proposed test-time scaling methods are effective in selecting better reasoning during model inference.}
  \label{fig:infer_demo}
\end{figure*}

\section{Further Experiments on the Effectiveness of CoT}

To further demonstrate the effectiveness of the CoT design in \ours, we supplement more experiments in Table~\ref{tab:analysis_cot}.

First, we aim to verify the effectiveness of the CoT design and \texttt{\ours-100k}. To this end, we use the prompts designed for creating the \texttt{\ours-100k} dataset to prompt GPT-4o and Llama-3.2-Vision to generate responses. GPT-4o performs much better under our structured CoT prompting. While the exact CoT design used during GPT-4o’s training is unknown, the results suggest that our structured CoT design is more effective, leading to a clear performance gain. Therefore, we are not simply replicating its training strategy, nor are we distilling existing capabilities of GPT-4o. Instead, we reveal the effectiveness of the CoT approach itself. However, structured CoT prompting does not improve Llama’s performance, indicating that prompting alone is insufficient without training. In contrast, our SFT on structured CoT data improves Llama-3.2-Vision, demonstrating the effectiveness of our method.

Second, we want to show that the improvement of \ours comes from CoT, rather than dense supervision from GPT-generated data. Therefore, we split our data for multi-task training of a captioner, summarizer, etc., without using CoT but still injecting GPT-4o supervision. The performance is significantly worse, suggesting that denser supervision from GPT-4o is not the reason for improvement.

Finally, since we point out that the logical order of SUMMARY, CAPTION, REASONING, and CONCLUSION is one of the key reasons for the effectiveness of our structured CoT, we need to show that only when the four stages follow a natural reasoning process does the reasoning become effective. Thus, we train with shuffled stage orders, which leads to almost no improvement, indicating that the proper order is crucial for response quality.

\section{Hyperparameters in Figure~\ref{fig:scaling}}

The results in Figure~\ref{fig:scaling} were obtained by changing scaling parameters (the number of candidates / retracing iterations) for three algorithms. For \textbf{Best of N}, $N$ values for the four experimental points are 1, 3, 4, and 8. For \textbf{Stage-wise Beam Search}, the number of candidates for the four points are 1, 4, 6, and 19. For \textbf{SWIRES}, the number of retracing iterations for the three points are 0, 1, and 3.

\section{Comparison Before and After Using Test Time Scaling}
As shown in Figure \ref{fig:infer_demo}, we demonstrate the performance of \ours\ with and without test-time scaling. From the figure, it can be observed that test-time scaling effectively corrects the errors made by the model during generation.

\section{Further Comparisons of Generated Content}
\begin{figure*}[h]
  \centering
  \vspace{-1em}
  \includegraphics[width=0.85\linewidth]{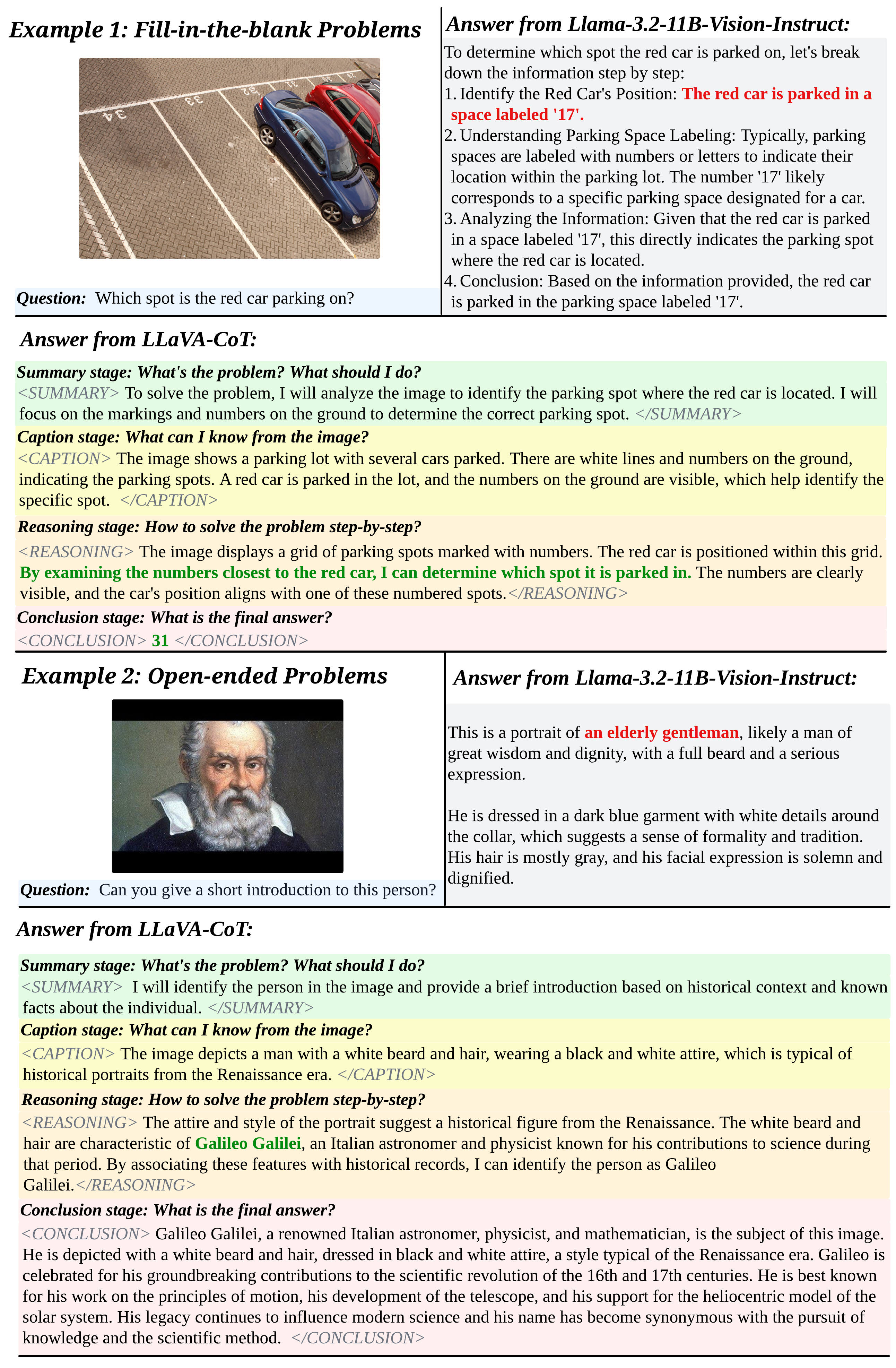}
  \vspace{-1em}
    \caption{Comparison of responses from the base model and \ours \ on fill-in-the-blank and open-ended problems.}
  \label{appfig:short}
\end{figure*}

The main paper provides examples focusing on reasoning and science-based multiple-choice questions. In this section, we extend the comparison to fill-in-the-blank and open-ended problems, further demonstrating the effectiveness of \ours. As illustrated in Figure \ref{appfig:short}, Llama-3.2-11B-Vision-Instruct frequently encounters issues when responding to fill-in-the-blank and open-ended questions, such as a lack of specificity or systematic reasoning. These shortcomings often result in factual inaccuracies or responses that are overly vague and fail to address the core aspects of the question.

\section{Limitations}

\ours \ also has certain limitations. Sometimes, \ours \ gets lost during retracing or starts hallucinating in order to reach an answer. By analyzing failure cases, we found that in rare cases, the input image may be overly complex and exceed the model’s visual understanding capabilities. As a result, even after retracing, the model may still fail to produce the correct answer.

\end{document}